\documentclass[10pt,twocolumn,letterpaper]{article}

\usepackage[pagenumbers]{cvpr} 

\usepackage[dvipsnames]{xcolor}

\definecolor{cvprblue}{rgb}{0.21,0.49,0.74}
\usepackage[pagebackref,breaklinks,colorlinks,citecolor=cvprblue]{hyperref}
\usepackage{graphicx}
\usepackage{multirow}
\usepackage[accsupp]{axessibility} 
\usepackage{adjustbox}
\usepackage{tabularx}

\def\confName{CVPR}
\def\confYear{2024}

\title{TrafficVLM: A Controllable Visual Language Model for Traffic Video Captioning}
\author{Quang Minh Dinh\textsuperscript{1} \qquad Minh Khoi Ho\textsuperscript{2} \qquad Anh Quan Dang\textsuperscript{2} \qquad Hung Phong Tran\textsuperscript{2} \\
\textsuperscript{1} Simon Fraser University, BC, Canada\\
\textsuperscript{2} Hanoi University of Science and Technology, Hanoi, Vietnam\\
{\tt\small qmd@sfu.ca}\\
{\tt\small \{khoi.hm204917, quan.da203909, phong.tnh200465\}@sis.hust.edu.vn}
}

\begin{document}
\maketitle
\begin{abstract}
Traffic video description and analysis have received much attention recently due to the growing demand for efficient and reliable urban surveillance systems. Most existing methods only focus on locating traffic event segments, which severely lack descriptive details related to the behaviour and context of all the subjects of interest in the events. In this paper, we present TrafficVLM, a novel multi-modal dense video captioning model for vehicle ego camera view. TrafficVLM models traffic video events at different levels of analysis, both spatially and temporally, and generates long fine-grained descriptions for the vehicle and pedestrian at different phases of the event. We also propose a conditional component for TrafficVLM to control the generation outputs and a multi-task fine-tuning paradigm to enhance TrafficVLM's learning capability. Experiments show that TrafficVLM performs well on both vehicle and overhead camera views. Our solution achieved outstanding results in Track 2 of the AI City Challenge 2024, ranking us third in the challenge standings. Our code is publicly available at \url{https://github.com/quangminhdinh/TrafficVLM}.
\end{abstract}    
\section{Introduction}
\label{sec:intro}
The recent advancements in dense video captioning models, which can precisely localize and describe incidents within a continuous video stream, have brought new opportunities and challenges to the field. This capability is particularly crucial in complex urban environments where the dynamic interactions between pedestrians, vehicles, and other elements can lead to accidents. As urban areas continue to grow and traffic becomes denser, the ability to automatically and accurately identify and describe accident scenarios from multiple perspectives becomes essential. This not only aids in immediate response efforts but also informs the development of safer, more intelligent transportation solutions. 

Most traditional deep learning systems for this particular domain virtually utilize the supervising training approach to predict frames or localize the incident segments \cite{7564410, futureFramesAnomaly, 9712446}. While models of this kind may capture the nuanced details of the traffic interactions, they lack the ability to explain their predictions, making it difficult to analyze causes, predict outcomes, and develop preventative measures. In response to this challenge, the 8\textsuperscript{th} AI City Challenge \cite{Shuo24AIC24} introduces the Traffic Safety Description and Analysis task, which involves detailed video captioning of traffic safety scenarios for both vehicles and pedestrians, given the videos from multiple static overhead cameras or moving vehicle ego cameras.

In this work, we introduce TrafficVLM, which leverages the advancements of multi-modality dense video captioning models, specifically adapted to the traffic domain. TrafficVLM extracts different layers of visual features from the vehicle camera frames to locate different phases of the traffic events and then provide detailed descriptions for different targets. Our contributions can be summarized as follows:

\begin{itemize}
  \item We reformulate the multi-phase Traffic Safety Description and Analysis task as a temporal localization and dense video captioning task with a single sequence as the output and introduce TrafficVLM, a video language model that is adapted specially to this task and domain.
  \item In accordance with our new fine-tuning objective, we propose a method to model the video features at different levels, enabling our model to effectively capture the fine-grained visual details, both spatially and temporally.
  \item We make use of the availability of captions for different targets in the dataset to devise a multi-task fine-tuning paradigm, allowing TrafficVLM to effectively learn the alignments between the video and textual features for all phases.
  \item We achieved the third rank on the blind test set of the AI City Challenge 2024 Track 2 with high results, which show the competitiveness of our solution.
\end{itemize}

\section{Related Works}
\label{sec:related}
\subsection{Traffic Accident Detection with Descriptions}
Traffic accident detection aims to identify accidental traffic incidents during driving, such as collisions among vehicles and objects and loss of control. Early methods for this task \cite{7564410} involve using a manual feature extraction process and simply utilizing a Bayesian model to detect traffic incidents, which lacks the ability to generalize and is sensitive to the extracted features. Further advancements in deep neural networks allow researchers to deploy deep learning approaches to reconstruct and predict errors in video frames to detect traffic accidents \cite{futureFramesAnomaly, hasan2016learning, chong2017abnormal}. In recent years, to counter the problem of hectic backgrounds, researchers have applied a two-stage process. Initially, architectures such as Mask-RCNN \cite{he2018mask}, FlowNet \cite{IMKDB17}, DeepSort \cite{Wojke2017simple} or ORBSLAM \cite{murAcceptedTRO2015} are used, for instance by Yao \etal \cite{9712446} to extract visual features such as bounding boxes, optical flow, tracking ids, and ego-motion correspondingly. Then, a detection process is applied to these features for the final classification.

On the other hand, the latest applications in deep learning revolve around the power of multi-modality models. For instance, Liang \etal \cite{liang2024textdriven} exploit the language features, which are already aligned with visual features by extensive training of CLIP \cite{radford2021learning}, as supervised signals while capturing the dynamic changes of driving scenes in the high temporal domain. A few other methods even approach the problem from a generative perspective, such as TRIVIA \cite{qasemi2023trafficdomain}, which infuses traffic-domain knowledge into a large video-language model to elevate the combined capabilities derived from having been multi-modality pre-trained. 
 
\subsection{Video Visual Language Models}
Together with the advancements of image-text pre-training models \cite{wang2021minivlm, yu2022coca, zhang2021vinvl, wang2022simvlm, wang2021ufo, wang2022git, li2022blip, huang2020pixelbert, dou2022empirical, alayrac2022flamingo, radford2021learning}, recent works are moving toward video-text pre-training, with some of the representatives are \cite{xu2021videoclip, yan2023videococa, zhao2024videoprism, li2024unmasked, wang2022internvideo}. Similar to image-text pre-training, while these models are excellent at capturing global semantic understanding of the videos, they still struggle with temporal localization and are not suitable to use off-the-shelf in a generative manner. 
\subsection{Dense Video Captioning and Localization}
Dense video captioning takes a further step ahead to understand events in the video as it uses timestamp information. Solutions for this task often interest in capturing all the details in a video that describe major events that happen inside it. The general flow for dense video captioning consists of three steps: (1) Video Feature Extraction, (2) Temporal Event Localization, and (3) Dense Caption Generation. Video feature extraction commonly uses strong video visual language models mentioned above, while traditional approaches for temporal event localization can be categorized into two types: proposal-based \cite{zhang2019man, yuan2019semantic, chen-etal-2018-temporally} and proposal-free methods \cite{zhang2020spanbased, yuan2018talk, ghosh2019excl}. Proposal-based techniques generate candidate proposals before ranking them based on relevance. In contrast, proposal-free methods directly predict the start and end boundaries of the target moment. Early approaches often complete the dense caption generation task by adding dedicated transformer models \cite{yamazaki2023vltint, yamazaki2022vlcap, yan2021dvcflow}, while more recent approaches \cite{Yang2023Vid2SeqLP, yang2023vidchapters, huang2023vtimellm, wang2024omnivid} in this field have
witnessed a shift towards joint training of captioning and
localization modules. For instance, Vid2Seq \cite{Yang2023Vid2SeqLP} enhances a language model by incorporating specific time tokens, enabling the model to generate event boundaries and textual descriptions within the unified output sequence. Furthermore, VTimeLLM \cite{huang2023vtimellm} leverages the power of a large language model to enable natural language interaction with humans, while presenting excellent generalization in video understanding and event localization.

\section{Method}
\label{sec:method}

\begin{figure*}[t]
    \centering
   \includegraphics[width=\textwidth]{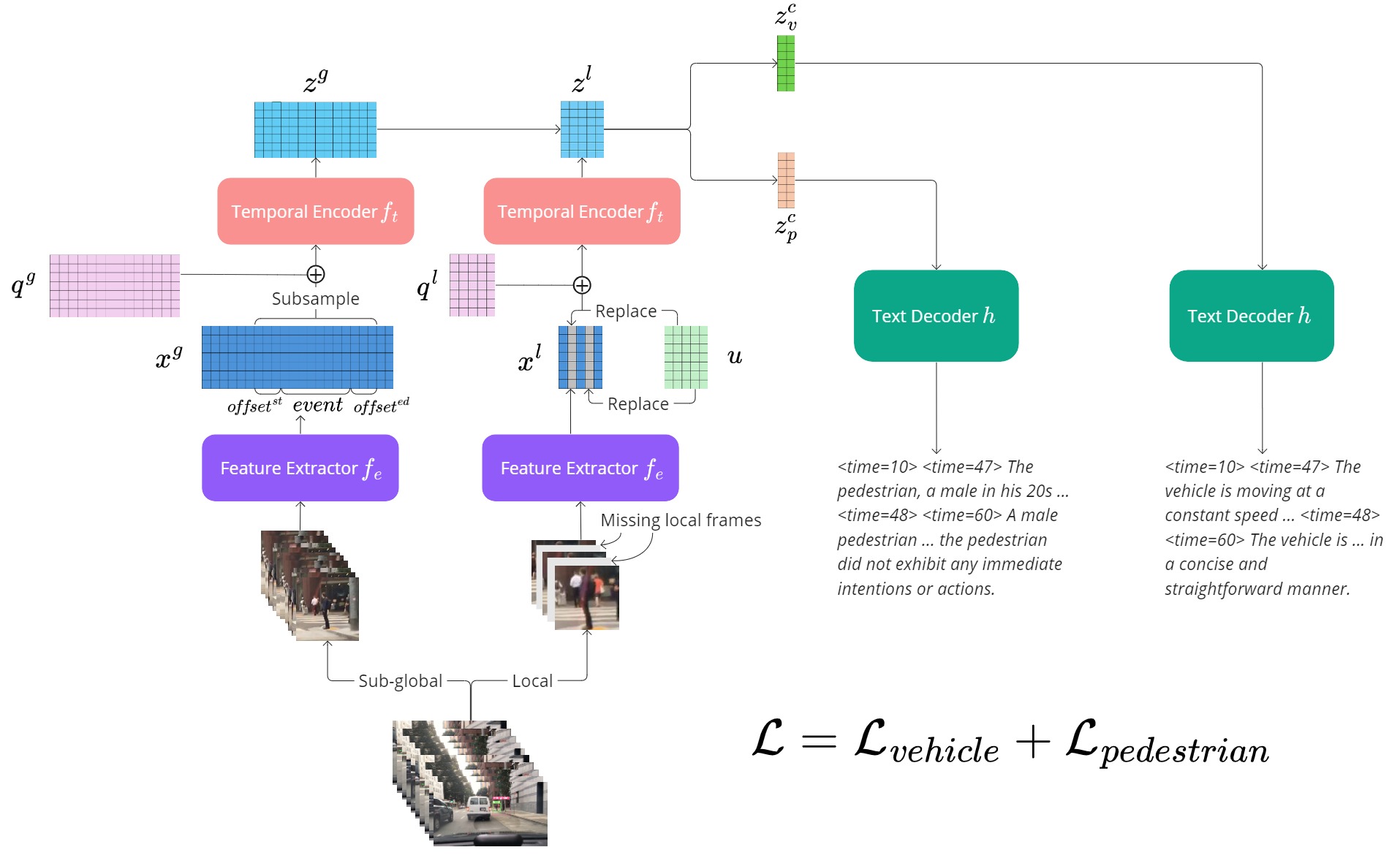}
   \caption{Overview of our method. First the sub-global and local frame sequences are extracted from the vehicle camera video. Some local frames might be missing, depending on the availability of the bounding boxes in the event segment. A visual feature extractor $f_e$ is then applied to both of them to get the two visual embeddings $x^g$ and parts of $x^l$. The sub-global embedding is trimmed to the event segment and subsampled to create the sub-global feature $\tilde{x^g}$. Feature vectors in the learnable local tensor $u$ are added to the local embedding as the embeddings for the missing phases. Positional embeddings are then applied to both visual features, followed by the temporal encoder $f_t$. The final embeddings $z^g$ and $z^l$ are concatenated with the conditional embedding $z^c_v$ for \textit{vehicle} or $z^c_p$ for \textit{pedestrian} to control the generation output. The text decoder $h$ receives the concatenated embedding as the input and autoregressively generates the output sequence. For fine-tuning, the final loss is calculated by combining the losses for generating the \textit{vehicle} and \textit{pedestrian} output sequences.
   } 
   \label{fig:arch}
\end{figure*}

Traffic Safety Description and Analysis is a challenging task which involves the long fine-grained captioning of different continuous phases of traffic safety scenarios for multiple targets, given the camera video, the timestamp of each phase, and the target bounding box information for a number of frames. Specifically, for each phase, the goal is to describe the surrounding context, attention, location, and behaviour of the pedestrian and vehicle in detail. In this section, we present TrafficVLM, a video language model that involves fine-tuning a temporal transformer encoder and the decoder of a large language model. Following Vid2Seq \cite{Yang2023Vid2SeqLP}, we reformulate this task as a temporal localization and dense video captioning task wherein the model learns to predict both the event boundaries and the descriptions for a target as a single sequence of tokens. In \cref{sec:arch}, we introduce a controllable component which allows TrafficVLM to generate multiple captions for different targets. We also show how to make use of different target captions to enhance our fine-tuning paradigm in \cref{sec:finetune}.

\textbf{Problem Formulation.} Given a vehicle camera video $V\in\mathbb{R}^{T \times H \times W \times C}$ with $T$ frames, the event boundary sequence $s=\{(start_i, end_i)\}_{i=1}^P$ with $P$ phases, and a list of pedestrian bounding boxes $b=\{(x_i^{st},x_i^{ed}, y_i^{st}, y_i^{ed})\}_{i=1}^n$ for $n<T$ frames, the goal is to generate two sequences $\tilde{v}=\{\tilde{v_i}\}_{i=1}^{L^v}$ for \textit{vehicle} and $\tilde{p}=\{\tilde{p_i}\}_{i=1}^{L^p}$ for \textit{pedestrian} that contain both the temporal information and the textual descriptions for all $P$ phases. The final $P$ captions of both targets, $v=\{v_i\}_{i=1}^P$ and $p=\{p_i\}_{i=1}^P$, can be decoded and extracted directly from $\tilde{v}$ and $\tilde{p}$.

\subsection{Visual Feature Extraction}
\label{sec:feature}

\textbf{Main Feature.} To remove redundant information, we extract our main video feature at a sub-global level by cropping each frame equally to a target segment that contains all local pedestrian features across $T$ frames. Explicitly, we select a segment $(x_{min}^{st},x_{max}^{ed}, y_{min}^{st}, y_{max}^{ed})$ from the bounding box information $b$ and extend the shorter dimension to length
\begin{equation}
  \tilde{W}=\max(y_{max}^{ed}-y_{min}^{st},x_{max}^{ed}-x_{min}^{st})
  \label{eq:crop_size}
\end{equation}
to make the selected segment square. We crop all $T$ frames to this area, which results in $\tilde{V}\in\mathbb{R}^{T \times \tilde{W} \times \tilde{W} \times C}$. Our visual feature extractor, which is denoted as $f_e$, is a frozen CLIP ViT-L/14 \cite{radford2021learning} that processes at resolution $224\times224$ pixels. We resize each cropped frame to the target resolution and encode it independently to get the visual embedding for the entire video:
\begin{equation}
  x^g=\{f_e(\tilde{V_i}) | \tilde{V_i}\in\tilde{V}\}\in\mathbb{R}^{T \times d},
  \label{eq:encode_glob}
\end{equation}
with $d$ being the dimension of the embedding.

We continue to use the event boundary sequence $s$ to trim the video feature to the segment that is related to the traffic event. To diversify the segment duration and the phase timestamps for training, we randomly select two offset durations $o\hspace{-0.1em}f\hspace{-0.1em}f\hspace{-0.1em}set^{st}$ and $o\hspace{-0.1em}f\hspace{-0.1em}f\hspace{-0.1em}set^{ed}$ between 0s and 5s at the video extracted frame rate and change the segment start and end frame accordingly to increase the duration. Specifically, the video feature $x^g$ is trimmed to
\begin{equation}
    \begin{split}
  (\max(0,start_1 - o\hspace{-0.1em}f\hspace{-0.1em}f\hspace{-0.1em}set^{st}), \\
  \min(end_P + o\hspace{-0.1em}f\hspace{-0.1em}f\hspace{-0.1em}set^{ed}, T)).
  \end{split}
  \label{eq:trim_main}
\end{equation}
We subsample the trimmed feature to a smaller frame rate and then continue to subsample or zero-pad it to $F$ frames. The final feature $\tilde{x^g}\in\mathbb{R}^{F \times d}$ is used as the main video feature for training.

\textbf{Local Feature.} For each phase $P_i\in P$, we randomly select a single pedestrian bounding box $b_{P_i}\in b$ to construct our local features. Similar to how we process our main feature, we square crop each frame corresponding to bounding box $b_{P_i}$ to the segment that is close to the bounding box and resize it to $224\times224$ pixels, represented as $l_i\in\mathbb{R}^{224\times224\times C}$. The cropped segment is then separately encoded with our visual extractor to get:
\begin{equation}
  x^l_i=f_e(l_i) \in\mathbb{R}^d.
  \label{eq:encode_local_single}
\end{equation}

As illustrated in \cref{fig:arch}, there can be several scenarios where there is no bounding box $b_{P_j}$ in relation to phase $P_j\in P$. To address this issue, we add a learnable tensor $u=\{u_i\}_{i=1}^P\in\mathbb{R}^{P\times d}$, and replace the missing visual feature $x^l_j$ with the learnable embedding $u_j$ at the corresponding phase. All phase visual features are aggregated to form the final local embedding $x^l=\{x^l_i\}_{i=1}^P\in\mathbb{R}^{P\times d}$.

In practice, we extract all raw CLIP features $x^g$ and $x^l$ prior to training to save computation time and resources. Details about the selection of features will be explained in \cref{sec:ablation}.

\subsection{Architecture}
\label{sec:arch}

\textbf{Temporal Visual Encoder.} We model the temporal dynamics for the frames of both the sub-global feature $\tilde{x^g}$ and the local feature $x^l$ using two learnable positional embeddings $q^g\in\mathbb{R}^{F\times d}$ and $q^l\in\mathbb{R}^{P\times d}$, and a vision transformer \cite{dosovitskiy2021an} $f_t$, which results in the final visual embeddings $z^g$ and $z^l$:
\begin{equation}
  z^g=f_t(\tilde{x^g}+q^g) \in\mathbb{R}^{F\times d}
  \label{eq:temporal_glob}
\end{equation}
\begin{equation}
  z^l=f_t(x^l+q^l) \in\mathbb{R}^{P\times d}
  \label{eq:temporal_local}
\end{equation}

\textbf{Generation Output Control.} We make use of how the transformer decoder uses the multi-head attention layer to attend to the input of the encoders and design the conditional module in a way that allows the model to learn the condition itself. Simply, for both targets \textit{vehicle} and \textit{pedestrian}, we add the additional learnable conditional embeddings $z^c_v\in\mathbb{R}^{k\times d}$ and $z^c_p\in\mathbb{R}^{k\times d}$, where $k$ is the extended dimension of the embeddings.

\textbf{Text Decoder.} To generate the output sequences, we employ T5-Base \cite{Raffel2019ExploringTL} as our transformer decoder $h$. The output sequences contain both the pseudo-timestamps and the textual descriptions for all phases:
\begin{equation}
  \tilde{v}=\{\tilde{v_i}\}_{i=1}^{L^v}=h(z^g,z^l,z^c_v) 
  \label{eq:out_vehicle}
\end{equation}
\begin{equation}
  \tilde{p}=\{\tilde{p_i}\}_{i=1}^{L^p}=h(z^g,z^l,z^c_p),
  \label{eq:out_pedestrian}
\end{equation}
where $L^v$ and $L^p$ are the lengths of the \textit{vehicle} and \textit{pedestrian} sequences correspondingly. As the main video features are trimmed to the target scenario segments, we do not include the temporal boundaries information in the inputs to the decoder to force the model to learn the temporal alignment between the visual features of each phase and the matching textual description. Details about the output sequences will be given in the following parts.

\textbf{Time Tokenization.} As T5-Base was selected for our text decoder, we also use the T5 tokenizer, which is based on the SentencePiece tokenizer \cite{Kudo2018SentencePieceAS} and is publicly available on the HuggingFace library \cite{Wolf2019HuggingFacesTS}. Following Vid2Seq, we extend the tokenizer by adding $N=100$ additional time tokens, which represent the relative timestamps in each video segment.

\textbf{Ouput Sequence Construction.} This part explains how we construct the two output sequences to be used as groundtruths during training and their format. We first adjust the event boundary sequence $s$ to the new start time $start^n=\max(0,start_1 - o\hspace{-0.1em}f\hspace{-0.1em}f\hspace{-0.1em}set^{st})$ and end time $end^n=\min(end_P + o\hspace{-0.1em}f\hspace{-0.1em}f\hspace{-0.1em}set^{ed}, T)$ at \cref{eq:trim_main} to create the new event boundary sequence $s^g$:
\begin{equation}
\begin{split}
  s^g&=\{(start_i^g, end_i^g)\}_{i=1}^P\\
  &=\{(start_i-start^n, end_i-start^n)\}_{i=1}^P
\end{split}
  \label{eq:event_seq}
\end{equation}
The new duration $D$ of the video segment is obtained by:
\begin{equation}
  D=end^n-start^n
  \label{eq:duration}
\end{equation}
Each timestamp in the new event boundary sequence $s^g$ is then rescaled to an integer between $0$ and $N-1$ to construct the quantized event boundary sequence $s^q$:
\begin{equation}
  s^q=\left\{\left(\left\lfloor\frac{start_i^g \times N}{D}\right\rfloor, \left\lfloor\frac{end_i^g \times N}{D}\right\rfloor\right)\right\}_{i=1}^P
  \label{eq:quantize}
\end{equation}
We map each quantized timestamp to one of $N$ time tokens, the result sequence is denoted as $t$:
\begin{equation}
  t=\{(t_i^{st},t_i^{ed})\}_{i=1}^P
  \label{eq:time_tokens}
\end{equation}
For the \textit{vehicle} textual tokens, we first tokenize all phase descriptions of the caption sequences $v$ to obtain $v^t=\{v^t_i\}_{i=1}^P$, where $v^t_i=\{v^t_{i_j}\}_{j=1}^{L^v_i}$ is the textual token sequence for phase $i$. Similar to Vid2Seq, we construct a new sequence for each phase $i$ by concatenating the start time token $t_i^{st}$, the end time token $t_i^{ed}$, and all the textual tokens $v^t_{i_j}$ in $v^t_i$. We concatenate all such sequences in increasing order of the start time and add a $BOS$ token at the beginning and a $EOS$ token at the end of the sequence. The result is the final output sequence $\tilde{v}$:
\begin{equation}
\begin{split}
  \tilde{v}&=\{\tilde{v_i}\}_{i=1}^{L^v}\\
  &=[BOS, t_1^{st}, t_1^{ed}, v^t_{1_1}, ..., v^t_{1_{L^v_1}},t_2^{st},...,EOS]
\end{split}
  \label{eq:out_vehicle}
\end{equation}
The output sequence $\tilde{p}$ is obtained in the same manner:
\begin{equation}
\begin{split}
  \tilde{p}&=\{\tilde{p_i}\}_{i=1}^{L^p}\\
  &=[BOS, t_1^{st}, t_1^{ed}, p^t_{1_1}, ..., p^t_{1_{L^p_1}},t_2^{st},...,EOS]
\end{split}
  \label{eq:out_pedestrian}
\end{equation}

\textbf{Phase Captions Reconstruction.} During inference, the $P$ captions for both \textit{vehicle} and \textit{pedestrian} are recovered by using the tokenizer to decode $\tilde{v}$ and $\tilde{p}$, and regular expression to extract each of the captions from the combined sequences.

\begin{figure}
    \centering
    \begin{subfigure}{0.45\textwidth}
        \centering
        \includegraphics[width=\textwidth]{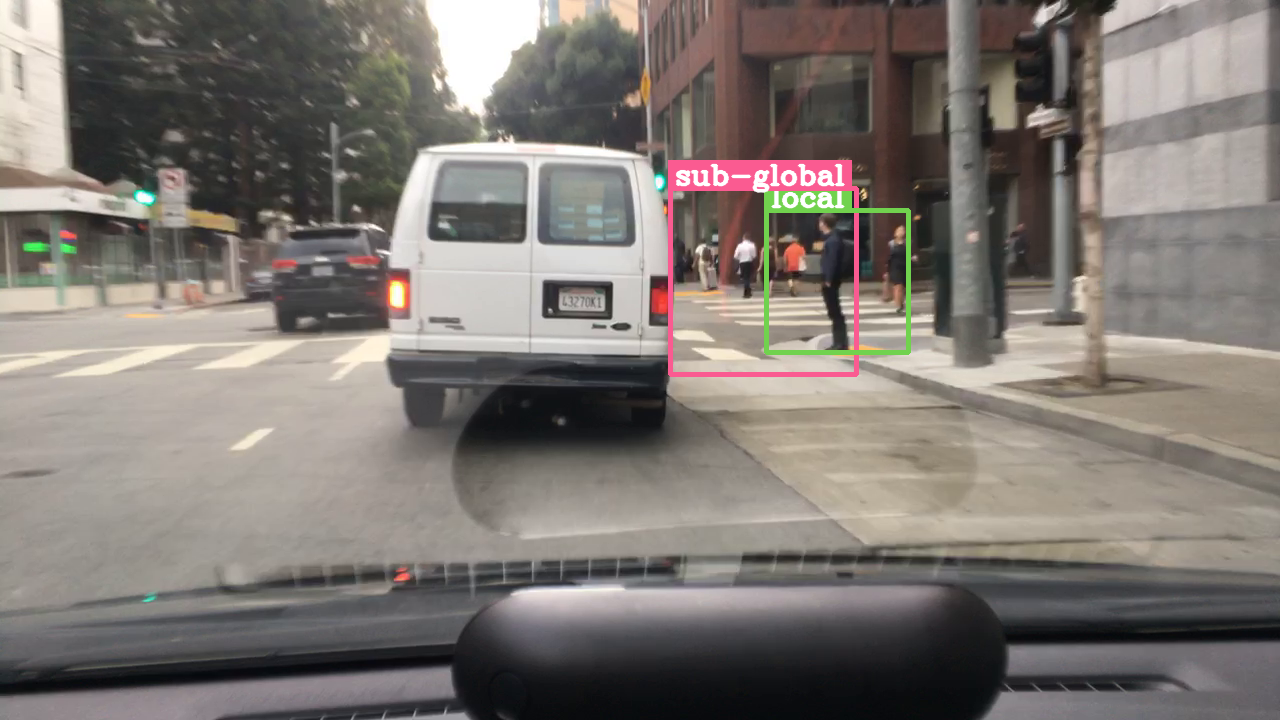}
        \caption{Vehicle view.}
        \label{fig:vehicle-view}
    \end{subfigure}
    
    \vspace{2mm}
    
    \begin{subfigure}{0.45\textwidth}
        \centering
        \includegraphics[width=\textwidth]{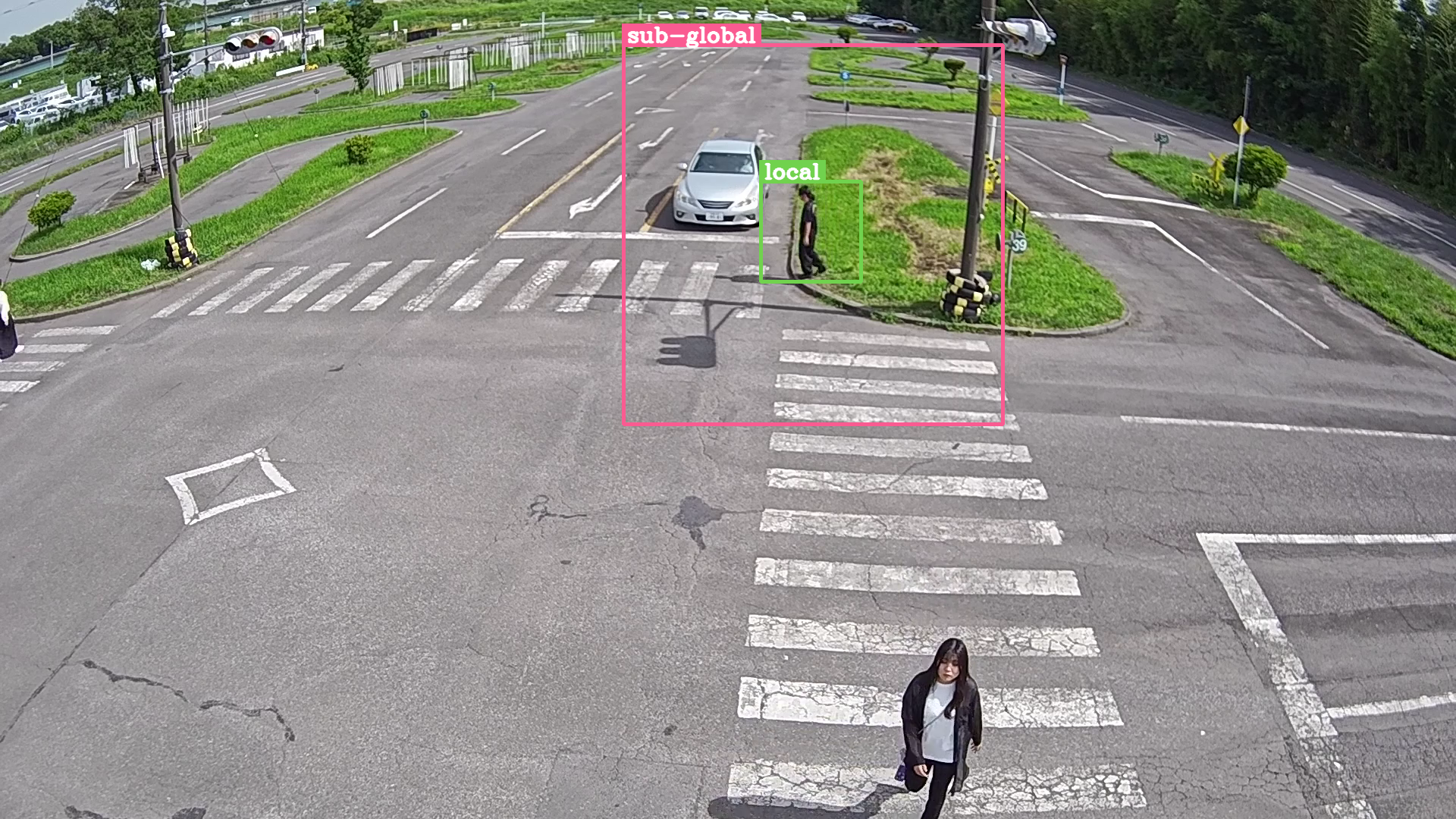}
        \caption{Overhead view.}
        \label{fig:overhead-view}
    \end{subfigure}
    \caption{Sub-global and local segments for vehicle and overhead camera views.}
    \label{fig:views}
\end{figure}

\subsection{Fine-tuning}
\label{sec:finetune}

To leverage the two different sets of captions provided by the dataset and let the model learn both the spatial and temporal alignments between the video and textual features better, we directly use \textit{vehicle captioning} and \textit{pedestrian captioning} as our two fine-tuning tasks. For both tasks, we apply the standard maximum likelihood objective, with teacher forcing \cite{teacherForcing} and a cross-entropy loss:
\begin{equation}
  \mathcal{L}_{\theta}^c(z^g,z^l,z^c,y)=-\mathbb{E}\left[\sum_{i=1}^{L-1} \log p_\theta\left(y_{i+1} \mid z^g,z^l,z^c,y_{<i}\right)\right],
  \label{eq:loss_ce}
\end{equation}
where $z^g$ is the sub-global video feature, $z^l$ is the local video feature, $z^c$ is the conditional embedding, $y$ is the target output sequence, and $L$ is the length of $y$.

The final loss is the combination of the losses of both tasks:
\begin{equation}
  \mathcal{L}_{\theta}=\mathcal{L}_{\theta}^c(z^g,z^l,z^c_v,\tilde{v})+\mathcal{L}_{\theta}^{c}(z^g,z^l,z^c_p,\tilde{p})
  \label{eq:loss}
\end{equation}

\section{Experiments}
\label{sec:experiment}

\subsection{Experimental Setup}


\textbf{Datasets.} For fine-tuning and evaluation, we use the WTS dataset \cite{WTS2024}, which is introduced along with the Traffic Safety Description and Analysis track in the AI City Challenge 2024. The dataset contains 155 scenarios and 810 videos from both fixed overhead cameras and vehicle cameras, as illustrated in \cref{tab:wts_stats}. In addition, the dataset also provides 3402 vehicle camera videos extracted from the BDD100K dataset \cite{Yu2018BDD100KAD}. Each scenario has around 5 phases, for which there is an annotation file capturing the location, attention, behaviour, and context information of both the vehicle and pedestrian in detail. Target pedestrian bounding boxes are provided for a number of frames in each video, and target vehicle bounding boxes are also provided for the overhead videos.

As the vehicle camera videos take up the majority of the WTS dataset, our fine-tuning data pipeline mainly extracts the visual features from the vehicle videos. We still randomly add in some overhead camera videos, which occupy up to 10\% of the data pipeline to create diversity in the fine-tuning data. All of our experiments are evaluated on the main subset of the WTS validation set. We generate vehicle and pedestrian captions for all subsets in the WTS test set and submit them to the AI City Challenge portal to obtain the test results.

\begin{table}
  \centering
  \begin{tabular}{l | l | c | c}
    \multicolumn{1}{ c| }{Split} & \multicolumn{1}{ c| }{Subset} & No. scenes & No. videos \\
    \hline
    \multirow{3}{*}{Train} & Main & 97 & 425 \\
    & Normal trimmed & 70 & 114 \\
    & External &  \multicolumn{2}{ c }{2430}\\
    \hline
    \multirow{3}{*}{Valid} & Main & 48 & 210 \\
    & Normal trimmed & 34 & 61 \\
    & External &  \multicolumn{2}{ c }{972}\\
    \hline
    \multirow{3}{*}{Test} & Main & 49 & 224 \\
    & Normal trimmed & 35 & 65 \\
    & External &  \multicolumn{2}{ c }{375}\\
    \end{tabular}
  \caption{Statistics of the WTS dataset.}
  \label{tab:wts_stats}
\end{table}

\begin{table*} 
\centering
\setlength\tabcolsep{4.24pt}
\begin{tabular*}{\linewidth}{ c | c | c | c | c c c c | c c c c | c }
 \multirow{2}{*}{Global} & Sub- & \multirow{ 2}{*}{Local} & Phase &
 \multicolumn{4}{ c |}{Vehicle Captioning} & \multicolumn{4}{ c |}{Pedestrian Captioning} & \multirow{2}{*}{Score} \\
 & global & & Encoder & BLEU & ROUGE & METEOR & CIDEr & BLEU & ROUGE & METEOR & CIDEr & \\
 \hline
 & \checkmark & & & 0.395 & 0.584 & 0.478 & 0.607 & 0.302 & 0.37 & 0.424 & 0.323 & 33.08 \\
 \checkmark & & & & 0.426 & 0.591 & 0.489 & 0.316 & 0.302 & 0.376 & 0.42 & 0.512 & 33.55 \\
 \checkmark & \checkmark & \checkmark & \checkmark & 0.402 & 0.589 & 0.472 & 0.354 & 0.316 & 0.376 & 0.429 & \textbf{0.882} & 33.84 \\
 \checkmark & \checkmark & & & 0.432 & 0.594 & 0.495 & 0.703 & 0.313 & 0.375 & 0.425 & 0.257 & 34.12 \\
 & \checkmark & \checkmark & & 0.411 & \textbf{0.598} & 0.479 & 0.647 & \textbf{0.326} & 0.375 & 0.437 & 0.412 & 34.15 \\
 \checkmark & & \checkmark & \checkmark & 0.433 & 0.589 & 0.499 & 0.555 & 0.315 & \textbf{0.386} & \textbf{0.437} & 0.482 & 34.54 \\
 & \checkmark & \checkmark & \checkmark & \textbf{0.443} & 0.591 & \textbf{0.5} & \textbf{0.785} & 0.317 & 0.367 & 0.431 &  0.507 & \textbf{34.74} \\
\end{tabular*}
\caption{Ablation study results for our TrafficVLM variations on the WTS main validation set, vehicle view. \textit{Phase Encoder} denotes temporal modelling for the local feature. The final score used for selecting the models is calculated in the same manner as the evaluation score of the 8\textsuperscript{th} AI City Challenge Track 2.}
\label{tab:ablation_vehicle}
\end{table*}

\begin{table*} 
\centering
\setlength\tabcolsep{4.24pt}
\begin{tabular*}{\linewidth}{ c | c | c | c | c c c c | c c c c | c }
 \multirow{2}{*}{Global} & Sub- & \multirow{ 2}{*}{Local} & Phase &
 \multicolumn{4}{ c |}{Vehicle Captioning} & \multicolumn{4}{ c |}{Pedestrian Captioning} & \multirow{2}{*}{Score} \\
 & global & & Encoder & BLEU & ROUGE & METEOR & CIDEr & BLEU & ROUGE & METEOR & CIDEr & \\
 \hline
 \checkmark & & & & 0.394 & 0.564 & 0.467 & 0.384 & 0.297 & 0.373 & 0.416 & 0.171 & 32.08 \\
 & \checkmark & \checkmark & & 0.401 & 0.591 & 0.474 & 0.218 & \textbf{0.322} & 0.378 & 0.432 & 0.392 & 33.23 \\
 \checkmark & \checkmark & \checkmark & \checkmark & 0.413 & \textbf{0.597} & 0.479 & 0.252 & 0.306 & 0.377 & 0.433 & 0.345 & 33.31 \\
 & \checkmark & & & 0.395 & 0.588 & 0.474 & 0.601 & 0.314 & 0.374 & 0.43 & \textbf{0.661} & 33.76 \\
 \checkmark & \checkmark & & & 0.419 & 0.596 & 0.49 & 0.577 & 0.317 & \textbf{0.384} & 0.428 & 0.459 & 34.23 \\
 & \checkmark & \checkmark & \checkmark & \textbf{0.434} & 0.593 & 0.49 & 0.574 & 0.32 & 0.373 & 0.436 & 0.423 & 34.33 \\
 \checkmark & & \checkmark & \checkmark & 0.419 & 0.589 & \textbf{0.493} & \textbf{0.609} & 0.316 & 0.378 & \textbf{0.438} & 0.61 & \textbf{34.43} \\
\end{tabular*}
\caption{Ablation study results for our TrafficVLM variations on the WTS main validation set, overhead view.}
\label{tab:ablation_overhead}
\end{table*}

\textbf{Evaluation Metrics.} Following the evaluation setups of Track 2 of the 8\textsuperscript{th} AI City Challenge, we adopt BLEU-4 \cite{10.3115/1073083.1073135}, ROUGE-L \cite{lin-2004-rouge}, METEOR \cite{banerjee-lavie-2005-meteor}, and CIDEr \cite{vedantam2015cider} as our main evaluation metrics. BLEU and ROUGE are both standard metrics for assessing the quality of machine translation systems based on n-grams overlap that focus on precision and recall respectively. METEOR is an improvement of the BLEU score by using an $F_3$ measure of unigram similarity, alongside a penalty $p$ for displacements of words in generated sentences. CIDEr measures the similarity between the reference and the candidate sentences via the cosine similarity of their TF-IDF \cite{tfidf} weights. The final score which is used to rank the models is calculated as the combination of all 4 metrics:
\begin{equation}
    Score = \frac{1}{4}\times[100\times(B+M+R)+10\times C]
\end{equation}
where $B$, $M$, $R$, and $C$ stand for BLEU-4, METEOR, ROUGE-L, and CIDEr respectively.

\textbf{Implementation Details.} We initialize both the temporal visual encoder and the text decoder with checkpoint \textit{vid2seq\_htmchaptersvitt} from VidChapters \cite{yang2023vidchapters}, which is publicly available. The local temporal positional embedding $q^l$ is initialized by subsampling the sub-global positional embedding $q^g$, which is loaded from the VidChapters checkpoint. All videos are extracted at 30 FPS, and the main video features are subsampled to 3 FPS after the trimming. We set $F=100$ as the final number of frames for the sub-global feature. The text decoder is truncated or padded to $L^v=L^p=1024$ tokens. We apply the Adam optimizer \cite{Kingma2014AdamAM} with a learning rate $3\cdot 10^{-4}$, a cosine learning rate decay and a warm-up period. All of our models are fine-tuned with batch size 1 for 30 epochs on an NVIDIA RTX-3060 GPU for around 8 hours. We select the best checkpoint for each model based on the validation metrics.

\subsection{Ablation Studies}
\label{sec:ablation}

To study the effects of different modules and feature levels on the performance of our models, we mix and match them to create different variations of TrafficVLM, the results of which are shown in \cref{tab:ablation_vehicle} and \cref{tab:ablation_overhead}.

\textbf{Choice of features.} In addition to the sub-global and the local features used in the main model, we conduct some experiments with the features extracted at a global level by square cropping the entire camera frames centering on all the pedestrian bounding boxes. In \cref{tab:ablation_vehicle}, we present the ablation results for the Traffic Safety Description and Analysis task on the vehicle branch of the main WTS validation set. As our main concern is the vehicle camera view, we use the results in this table to rank the models. It can be seen that using a combination of two or more features is significantly better than just using one of them. In the experiments where the sub-global feature is directly used in comparison with its global counterpart (row 1 vs row 2 and row 7 vs row 6), the results of both are mostly comparable. The two best models use the global or sub-global feature alongside the local feature with temporal modelling. Of the two of them, the one with the sub-global feature (row 7) outperforms its counterpart in 6 out of 8 metrics. Unfortunately, any combination of both the global and sub-global features does not give a high result compared to other options.

\textbf{Performance on the Overhead Scenarios.} \cref{tab:ablation_overhead} shows the results of our experiments on the overhead branch of the main WTS validation set. Although overhead videos only take up to 10\% of our training set due to their availability compared to the vehicle videos, it is apparent that our models still perform relatively well in this scenario, as most of the numbers are close to those in \cref{tab:ablation_vehicle}. Two models with the global or sub-global feature used together with the temporal encoded local feature still perform the best in this scenario.

\textbf{Local Temporal Modelling.} As most scenarios only have around 5 phases, it is one of our designing concerns on whether to do temporal modelling for the local feature (which is denoted as \textit{Phase Encoder} in \cref{tab:ablation_vehicle} and \cref{tab:ablation_overhead}). We observe that in both scenarios, adding temporal modelling to the local feature significantly improves the model performance on most of the metrics (row 5 vs row 7 for \cref{tab:ablation_vehicle} and row 2 vs row 6 for \cref{tab:ablation_overhead}), which show the effectiveness of our design.

\subsection{Performance in the Challenge}

\begin{table}
  \centering
  \begin{tabular}{c | c | c}
    Rank & Team Name & Score \\
    \hline
    1 & AliOpenTrek & 33.4308 \\
    2 & AIO\_ISC & 32.8877 \\
    \textbf{3} & \textbf{Lighthouse} & \textbf{32.3006
}\\
    4 & VAI & 32.2778 \\
    5 & Santa Claude & 29.7838 \\
    6 & UCF-SST-NLP & 29.0084 \\
    7 & Monitor & 28.7485 \\
    8 & X & 27.7771 \\
    9 & HCMUS\_AGAIN & 22.7371 \\
  \end{tabular}
  \caption{Public leaderboard of the 8\textsuperscript{th} AI City Challenge Track 2.}
  \label{tab:leaderboard}
\end{table}

In Track 2 of the AI City Challenge 2024, we generate the result samples for the WTS internal blind test set (which contains both traffic scenarios and normal scenarios) with a model that uses only the sub-global feature. For the external blind set, we employ an ensemble method to combine that same model with the two best checkpoints of our main model using the generation confidence scores of the text decoder.

\cref{tab:leaderboard} presents the final public leaderboard of the 8\textsuperscript{th} AI City Challenge Track 2. Our team (with the team name Lighthouse) achieved third place with a final score of 32.3006. The result of our solution is only behind the top two teams by a small margin, which shows the competitiveness of TrafficVLM for the Traffic Safety Description and Analysis task.

\section{Conclusion}
\label{sec:conclusion}
In this paper, we present TrafficVLM, a visual language model tailored to perform dense video captioning for traffic incidents, as well as an effective fine-tuning paradigm leveraging the multi-target nature of the Traffic Safety Description and Analysis task. TrafficVLM achieved the third position in Track 2 of the AI City Challenge 2024 with an impressive score, demonstrating its effectiveness for the task. Besides dense video captioning and temporal localization, the current implementation of TrafficVLM could be extended to a variety of new tasks, including traffic video question answering, traffic video summarization, and other video understanding tasks.  We believe that future works can enhance TrafficVLM further by exploring the use of different large language models, such as Llama2 \cite{touvron2023llama} or Mistral \cite{jiang2023mistral} and employ different data augmentation strategies. The controllable design of TrafficVLM can be strengthened by using a text encoder to impose different conditions on the description generation.

{
    \small
    \bibliographystyle{ieeenat_fullname}
    \bibliography{main}
}


\end{document}